\documentclass{article}

\usepackage{PRIMEarxiv}

\usepackage[utf8]{inputenc} % allow utf-8 input
\usepackage[T1]{fontenc}    % use 8-bit T1 fonts
\usepackage{hyperref}       % hyperlinks
\usepackage{url}            % simple URL typesetting
\usepackage{booktabs}       % professional-quality tables
\usepackage{amsfonts}       % blackboard math symbols
\usepackage{nicefrac}       % compact symbols for 1/2, etc.
\usepackage{microtype}      % microtypography
\usepackage{lipsum}
\usepackage{fancyhdr}       % header
\usepackage{graphicx}       % graphics
\usepackage{amsmath}
\usepackage{amssymb}
\usepackage{booktabs}
\usepackage{subcaption}

\usepackage{algorithm}
\usepackage{algorithmic}
\graphicspath{{media/}}     % organize your images and other figures under media/ folder

\newcommand{\name}{ACE-EM}
\newcommand{\encoder}{$E_{IP}$}
\newcommand{\decoder}{$D_{PI}$}
\newcommand{\SI}{appendix}
\newcommand{\IPI}{IPI}
\newcommand{\PIP}{PIP}
\newcommand{\VGrid}{$\mathrm{VoxelGrid_{R}}$}
\newcommand{\FNet}{$\mathrm{FourierNet}$}

\newcommand{\citep}[1]{~\cite{#1}}
\newcommand{\cref}[1]{~\ref{#1}}

\begin{document}

\title{Boosted \textit{ab initio} Cryo-EM 3D Reconstruction with ACE-EM}

\author{Lin Yao$^{1}$, Ruihan Xu$^2$, Zhifeng Gao$^1$, Guolin Ke$^1$, and Yuhang Wang$^1$\\
\\
$^1$ DP Technology, Ltd., Beijing, China\\
$^2$ Peking University, Beijing, China\\
}

\maketitle

\begin{abstract}
    The central problem in cryo-electron microscopy (cryo-EM) is to recover the 3D structure from noisy 2D projection images which requires estimating the missing projection angles (poses). Recent methods attempted to solve the 3D reconstruction problem with the autoencoder architecture, which suffers from the latent vector space sampling problem and frequently produces suboptimal pose inferences and inferior 3D reconstructions.  Here we present an improved autoencoder architecture called ACE (Asymmetric Complementary autoEncoder), based on which we designed the ACE-EM method for cryo-EM 3D reconstructions. Compared to previous methods, ACE-EM reached higher pose space coverage within the same training time and boosted the reconstruction performance regardless of the choice of decoders. With this method, the Nyquist resolution (highest possible resolution) was reached for 3D reconstructions of both simulated and experimental cryo-EM datasets. Furthermore, ACE-EM is the only amortized inference method that reached the Nyquist resolution.
\end{abstract}

\vspace{-5pt}
\section{Introduction}
\vspace{-5pt}
Single-particle cryo-electron microscopy (cryo-EM) is one of the most important
structural biology techniques \citep{Frank_2017_Advances}. 
This technique can be divided into four stages: biological sample preparation,
electron microscopy image collection, 
3-dimensional (3D) reconstruction, 
and atomic structural model building.
3D reconstruction of molecular volumes from 2-dimensional (2D)
electron microscopy images is the most challenging and time-consuming step in cryo-EM data analysis. 
There are two major challenges in cryo-EM 3D reconstruction: 
unknown projection poses (orientations and positions) 
and low signal-to-noise ratio (SNR). 
During electron microscopy imaging, the 3D poses of the 
biological molecules in the sample can not be directly measured.
The SNR of a typical cryo-EM image is very low, which 
can vary from -10 dB to -20 dB (average around -10 dB) 
\citep{Bepler_2020_NatCommun}, making it extremely challenging to accurately estimate poses and perform the 3D reconstruction.

Currently, 
many machine-learning (ML)-based methods have been proposed for solving cryo-EM 3D reconstruction~\citep{2022Deep}, utilizing architectures like GAN~\citep{goodfellow2014generative} and auto-encoders~\citep{https://doi.org/10.1002/aic.690370209}.
Nevertheless, ML-based methods of cryo-EM 3D reconstruction
are still at an early stage.
The highest possible cryo-EM reconstruction resolution (2 pixels) 
at FSC threshold 0.5 has not been achieved by ML-based methods, 
even in simulated datasets without noise.
For experimental datasets like the 80S,
    some methods with amortized inference methods
    failed to reconstruct the object of certain size,
    or found the highest possible resolution (2 pixels) 
   of half-map FSC~\citep{ROSENTHAL2015135} at 0.143 not reachable~\citep{Levy_2022_ECCV}.
For widely utilized architecture auto-encoders, 
an image-to-pose encoder extracts the image-projection poses from 2D input cryo-EM images, while a pose-to-image decoder generates the images to match the inputs. 
However, as the poses are intermediate latent variables without supervised loss, the estimated poses can be inaccurate and easily trapped in local minima of the orientation space. These pose estimation errors lead to an inferior resolution in the 3D reconstruction output. 

To improve the pose estimation and 3D reconstruction quality, here we propose a new framework called \name{} (ACE for Asymmetric Complementary autoEncoder).
In particular, ACE-EM consists of two training tasks:
(1) Image-pose-image (IPI). The task is the same as previous work, which takes projection images as inputs, and outputs predicted images, by an image-to-pose encoder followed by a pose-to-image decoder.
(2) Pose-image-pose (PIP). The task explicitly learns the pose estimation in a self-supervised fashion, which takes randomly sampled poses as inputs, and outputs predicted poses, using the same encoder and decoder as in IPI but reversing their order.
The two tasks complement each other and achieve a more balanced training of the autoencoder parameters. \looseness=-1

Our main contributions are listed below.
\begin{itemize}
    \item As far as we know, ACE-EM is the first deep-learning model in cryo-EM reconstruction that enhances the pose estimation by the self-supervised PIP task. With better pose estimation, ACE-EM can converge much faster than previous methods, efficiently cover more pose spaces, and achieve better cryo-EM 3D reconstruction quality.
    \item ACE-EM can boost performance regardless of decoder types. 
    For example, some decoders, that failed in previous autoencoder architectures, can be successfully used in \name{}.
    \item Experimental results demonstrate that \name{} can perform well in both simulated and real-world experimental datasets. In particular,  \name{} outperformed all the baseline methods and reached 2 pixels, the Nyquist resolution (the highest possible resolution) at FSC threshold 0.5, 
    the Nyquist resolution of half-map FSC at 0.143 
 in the 80S experimental dataset,
    which is the only architecture that reached the Nyquist resolution with amortized inference methods. 

\end{itemize}

\vspace{-5pt}
\section{Related work}
\vspace{-5pt}
\paragraph{Overview of cryo-EM methods}
3D reconstruction and view synthesis is a popular field in computer vision. Many new methods have been emerged in recent years, such as
NeRF~\citep{DBLP:journals/corr/abs-2003-08934},
Plenoxels~\citep{DBLP:journals/corr/abs-2112-05131},
DirectVoxGo~\citep{https://doi.org/10.48550/arxiv.2206.05085},
BARF~\citep{DBLP:journals/corr/abs-2104-06405},
SCNeRF~\citep{DBLP:journals/corr/abs-2108-13826}, and
GNeRF~\citep{DBLP:journals/corr/abs-2108-13826}.
These methods can reconstruct 3D scenes from 
natural images with high signal-to-noise ratios (SNR).
In \textit{ab initio}
cryo-EM 3D reconstruction, the reconstruction problem is much harder 
due to the low-SNR nature of cryo-EM images and missing image
projection parameters. 
In addition to traditional cryo-EM reconstruction methods like RELION \citep{Scheres_2012_JSB}
and cryoSPARC \citep{Punjani_2017_NatMeth}, many
ML-based cryo-EM reconstruction methods have been developed in the past
few years.
Given only the 2D cryo-EM projection images without any labels, 
cryo-EM 3D reconstruction can be formulated as a self-supervised learning problem.
Two ML frameworks have been explored: GAN~\citep{goodfellow2014generative} and autoencoder~\citep{https://doi.org/10.1002/aic.690370209}.
CryoGAN \citep{Gupta_2021_IEEE} 
and Multi-CryoGAN \citep{Gupta_2020_ECCV} adopted the GAN framework.
In these methods, the generator contains the predicted volume information and generates projection images, while the discriminator adopts the generated and authentic images and discriminates the source of the images.
However, due to the limitations of GANs, 
these methods lack pose estimation and 3D reconstruction quality is inferior to other methods.
For cryo-EM, the images are fed to the encoder to generate a latent variable (usually representing the pose estimation), 
and the decoder produces the reconstructed images based on this variable.
The model is trained by reducing the loss of the original and reconstructed ones. 
After training, the 3D reconstruction density map can be obtained from the decoder.
Cryo-EM 3D reconstruction with variational autoencoder  (VAE) \citep{Ullrich_2019_PMLR}, 
CryoDRGN~\citep{Zhong_2019_ICLR,Zhong_2021_NatMeth,Zhong_2021_ICCV},
AtomVAE~\citep{Rosenbaum_2021_NIPS},
CryoPoseNet~\citep{Nashed_2021_ICCV},
and cryoAI~\citep{Levy_2022_ECCV}
adopt the autoencoder architecture.
They improve the performance of autoencoders in cryo-EM 3D reconstructions by using more powerful networks
for encoders and decoders, adding or modifying loss functions, modifying the latent variable designs
and other strategies.

\vspace{-8pt}
\paragraph{Pose estimation}
Pose estimation is concerned with 
estimating the missing
orientation and position (together known as pose) parameters
of the images based on the underlying 3D object. 
Existing pose estimation methods can be divided into two classes:
Per-image pose search methods and amortized inference methods.
The first class of methods performs a global pose search
for each projection image in the input dataset.
Traditional software,
like RELION \citep{Scheres_2012_JSB} 
and cryoSPARC \citep{Punjani_2017_NatMeth}, and some ML-based 
methods, 
inluding cryoDRGN-BNB \cite{Zhong_2019_ICLR} and 
CryoDRGN2\citep{Zhong_2021_ICCV},
fall into this category. 
The per-image pose search method is not an ML-based strategy,
To control the Pose estimation as an ML process,
the second class method, amortized inference,
is first proposed for cryo-EM 3D reconstruction 
by Ullrich \textit{et. al.}~\citep{Ullrich_2019_PMLR}.
It focuses on learning 
a parameterized function for mapping image $Y_i$ to its pose $\phi_i$, 
then it is widely accepted by autoencoder solutions as the encoder.
For autoencoders with amortized inference method,
they suffer from too many local minima in their 
probabilistic pose estimation framework.
Rosenbaum \textit{et. al.} attempted to overcome local-minima traps using variational autoencoder  (VAE) \citep{Kingma_2014_ICLR}.
CryoAI~\citep{Levy_2022_ECCV} extended CryoPoseNet~\citep{Nashed_2021_ICCV} work by introducing ``symmetry loss''
to facilitate pose learning by adding a 180$^\circ$-rotated input image as input.
In practice, we found these strategies are
insufficient for avoiding poses being trapped in local minima.
The proposed \name{} method also falls into this category of amortized inference. \looseness=-1

\vspace{-8pt}
\paragraph{Decoder or generator choices} 
In the view of cryo-EM 3D reconstruction, a decoder (in autoencoder) or a generator (in GAN) is a structure that contains the volume representation, which adopts the pose or other variables and outputs the predicted 2D projection images.
The underlying target 3D object can have either 
real-space or frequency-space representations in the space domain,
and either neural network type or voxel grid type for volume parameter representations~\citep{2022Deep}.
Traditional softwares represent
the reconstruction target volume as a
3D voxel grid of frequency space,
like Relion \citep{Scheres_2012_JSB} and
cryoSPARC \citep{Punjani_2017_NatMeth}.
Many recently developed deep-learning-based reconstruction methods mainly use two types of decoders (generators):
real-space voxel grid type
and frequency-space neural network type.
CryoDRGN~\citep{Zhong_2019_ICLR,Zhong_2021_NatMeth,Zhong_2021_ICCV} and CryoAI~\citep{Levy_2022_ECCV} uses the frequency-space neural network as their decoders.
Real-space voxel grid methods are also used by many algorithms, such as 
CryoGAN \citep{Gupta_2021_IEEE}, 
Multi-CryoGAN \citep{Gupta_2020_ECCV},
3DFlex \citep{Punjani_2021_bioRxiv}, 
and AtomVAE   \citep{Rosenbaum_2021_NIPS}.
CryoPoseNet \citep{Nashed_2021_ICCV} stores a real space voxel grid representation of 
the reconstruction target, but the projection was made in Fourier space. 
In this work, we show that the \name{} method is effective with both real-space voxel grid type
and frequency-space neural network type of decoders.
\vspace{-10pt}
\paragraph{Cyclic training} 
Our method uses encoder-decoder(E-D) and decoder-encoder(D-E) two orders for training in autoencoders, and this cyclic training is widely used in style transfer, like MUNIT~\citep{Huang_2018_ECCV}, DRIT~\citep{DBLP:journals/corr/abs-1905-01270} and CDD~\citep{DBLP:journals/corr/abs-1805-09730}. These methods differ from ours in 4 main aspects, architecture setting, data format, aims, and challenges.
For the architecture setting, MUNIT uses E-D and (E-)D-E two orders~\citep{Huang_2018_ECCV},
DRIT uses one task E-D-E-D to finish the cycle, while more complex tasks are used in CDD.
MUNIT and DRIT need two random images inputted together as a group for cross-domain translation~\citep{DBLP:journals/corr/abs-1905-01270,DBLP:journals/corr/abs-1805-09730}, CDD uses paired data for supervised learning~\citep{DBLP:journals/corr/abs-1805-09730}, but paired or grouped data is not required by \name{}.
Style transfer aims to decouple images' style and content, and our aim is to settle the poses in an accurate distribution.
Also, our methods face specific reconstruction challenges, like noise problems.

\vspace{-5pt}
\section{Approach}
\vspace{-4pt}
\label{Approach}

\subsection{Overview of \name{}}
\vspace{-6pt}
\name{} is an autoencoder-based model and is trained with two unsupervised learning tasks.
The autoencoder contains an image-to-pose encoder (\encoder{}) 
and a pose-to-image decoder (\decoder{}).
The \encoder{} takes projection images and outputs projection pose parameters. 
The \decoder{} can be viewed as a cryo-EM image projection physics simulator.
It takes the pose parameters and outputs projection images corresponding to the given poses, which are post-processed by applying CTF \citep{KONSTANTINIDIS201449}.
The first task of \name{} is 
image-pose-image(\IPI{}) task which follows the standard 
unsupervised autoencoder architecture.
With the pipeline of \encoder{} and \decoder{}, reconstructed images are generated corresponding to the given projection images.
The second task is the pose-image-pose (\PIP{}) task, a self-supervised learning, that uses the same encoder-decoder in IPI but in the reversed order. 
With the reverse pipeline, a corresponding pose is predicted from a random-selected pose, and the gap between the two poses is minimized in training.
The \IPI{} task and \PIP{} task can be trained simultaneously or 
in alternating steps. 
With either training strategy, 
the \encoder{} and \decoder{} parameters are shared 
between the two tasks. 
As a result, the two tasks of \name{} complement each other
and form a more balanced training of the \encoder{} and \decoder{}
parameters. \looseness=-1

\begin{figure*}[t]
		\centering
		\includegraphics[width=0.95\textwidth]{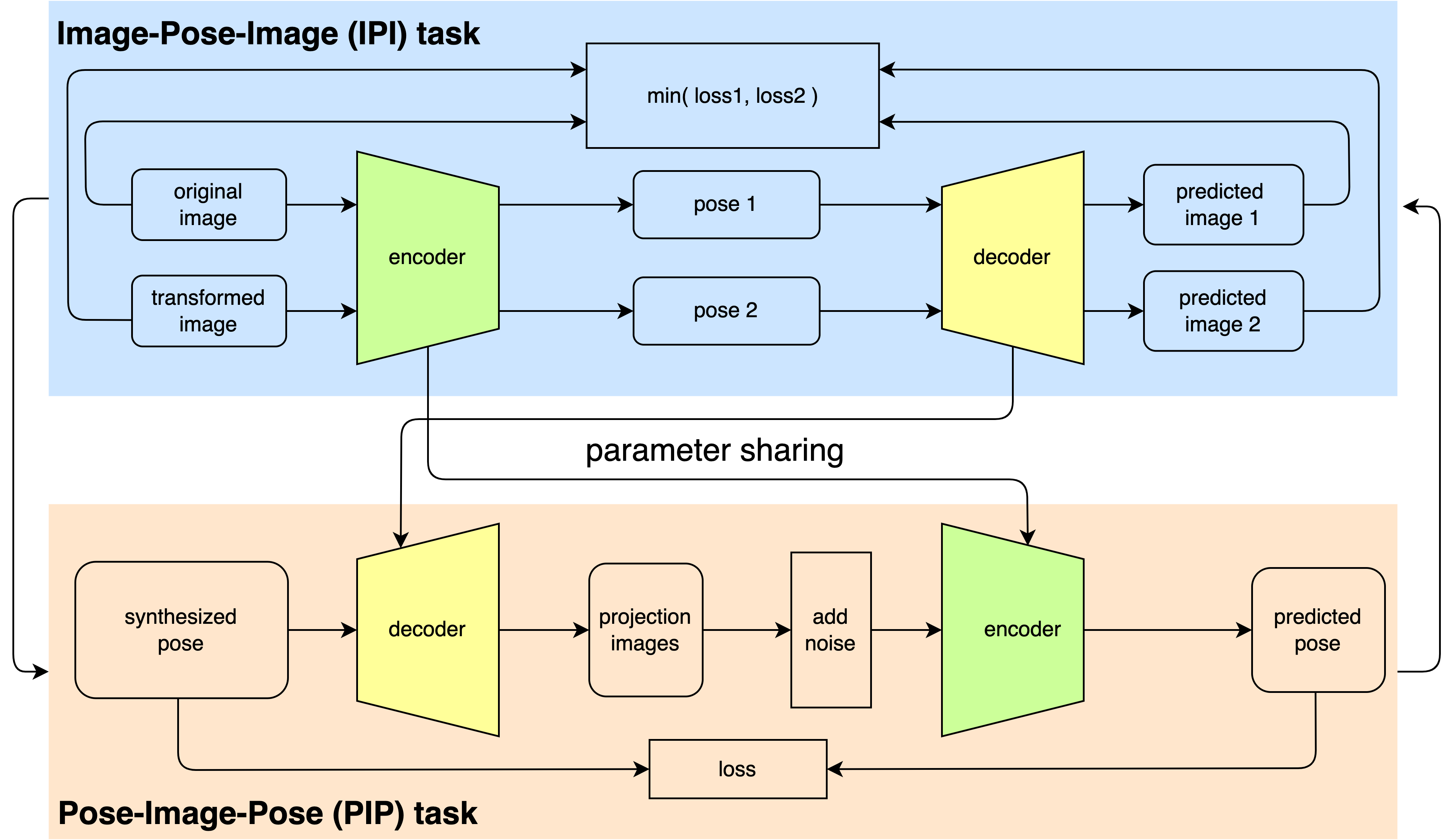}
        %\vspace{-10pt}
		\caption{The network architecture of ACE-EM. 
		The \IPI{} task is shown at the top highlighted in blue,
	    while the \PIP{} task is shown at the bottom highlighted in salmon.}
		\label{fig:arch}
\end{figure*}

\vspace{-8pt}
\subsection{Encoder \encoder{} and Decoder \decoder{}}
\vspace{-6pt}

The \encoder{} represents a function that maps an input image $Y_i$
to its corresponding projection pose parameters $(R_i, t_i)$.
$R_i \in SO(3) \subset \mathbb{R}^{3\times3}$ is a rotation matrix
for mapping a reference orientation to the projection orientation.
$t_i \in \mathbb{R}^2$ is the 2D translation vector to
account for the 2D image shift in the input image $Y_i$.
The \decoder{} takes pose $(R_i, t_i)$ and 
 predicts the projection image $Y_i^{pred}$.The trainable parameters of \decoder{} contain the cryo-EM volume in an explicit or implicit way, and the volume can be obtained from the \decoder{} after training.
CTF was applied to the output images
to create a more realistic projection image $Y_i^{pred}$.
 The details of the encoder and decoder structure can be found in \SI{} \ref{Encoder and decoder}.\looseness=-1
 
 \vspace{-10pt}
\begin{equation}
\small
E_{IP}: Y_i \mapsto (R_i, t_i)
\label{eqn:encoder-def1}
\end{equation}

\vspace{-10pt}

\begin{equation}
\small
    D_{PI}: (R_i, t_i) \mapsto Y_i^{pred}
    \label{eqn:decoder-def1}
\end{equation}
\vspace{-20pt}

\paragraph{Choices of decoders}

\name{} provides an architecture that can work with different
encoders and decoders. 
Cryo-EM 3D reconstruction can have either 
real-space or frequency-space representations in the space domain,
and either neural network type or voxel grid type for volume parameter representations.
To prove \name{} is universal and can boost the performances regardless of the decoder types, 
we tested a real-space voxel grid
decoder \VGrid{} which was 
used in cryoGAN~\citep{Gupta_2021_IEEE},
and partially used in the CryoPoseNet~\citep{Nashed_2021_ICCV},
and a frequency-space neural network decoder 
\FNet{} which was shown to outperform other similar methods
~\citep{Levy_2022_ECCV}. \looseness=-1

\vspace{-8pt}
\subsection{\IPI{} (Image-Pose-Image) task}
\vspace{-6pt}
The \IPI{} task follows the standard autoencoder architecture 
as shown in Figure~\ref{fig:arch}. Using the notations defined
earlier, the \IPI{} task can be formally defined as follows.

\vspace{-10pt}
\begin{equation}
\small
    \text{IPI}: Y_i \mapsto Y_i^{pred}
    \label{eqn:IPI-mapping}
\end{equation}

\vspace{-10pt}

\begin{equation}
\small
    f_{IPI}(Y_i) := (D_{PI} \circ E_{IP} )(Y_i)
    \label{eqn:IPI-def}
\end{equation}
\vspace{-15pt}

\paragraph{\IPI{} loss function}
Since both the input and output are images,
the objective of the \IPI{} task is to minimize
their differences by
mean squared error (MSE) loss function for each training batch of size $B$
and with an image edge length of $L$.
\begin{equation}
\small
    \mathcal{L}_{image} := \frac{1}{BL^2} \sum_{i=1}^{B} 
        \left\|Y_i - Y_i^{pred}\right \|_2
    \label{eqn:img-mse-loss}
\end{equation}
Cryo-EM 3D reconstruction is prone to spurious 2-fold planar mirror symmetry \citep{Levy_2022_ECCV}. 
A tentative solution is to use the ``symmetric loss" 
employed in cryoAI \citep{Levy_2022_ECCV}, where
$\Gamma_{\text{cryoAI}}$ is the cryo-AI autoencoder pipeline and
$R_\pi$ represents an in-plane rotation of angle $\pi$ operation
applied to the input image $Y_i$.
\begin{equation}
    \mathcal{L}_{\text{symm}}^{\text{cryoAI}} 
    := \frac{1}{BL^2}\sum_{i=1}^{B}
    \min(
        \left\|Y_i - \Gamma_{\text{cryoAI}}(Y_i)\right\|_2,
        \left\|R_\pi[Y_i] -  \Gamma_{\text{cryoAI}}(R_\pi[Y_i])\right\|_2
    )
    \label{eqn:symm-loss-cryoai}
\end{equation}
For a consideration of data augmentation, we have generalized the above symmetric loss to include 
in-plane image rotation by arbitrary angles and also included
mirror transformation.
``generalized symmetric loss" as below.
$A_1[Y_i]$ and $A_2[Y_i]$ are two different affine transformations
with random image rotations and/or mirror flipping.
We found that the generalized symmetry loss still works with the voxel-grid-based decoder \VGrid{}, but fails in \FNet{}, so it is only applied when using the \VGrid{}.
\begin{equation}
\small
    \mathcal{L}_{\text{symm}}^{G}
    := \frac{1}{BL^2}\sum_{i=1}^{B}
    \min(
        \left\|A_1[Y_i] -
            f_{IPI}(A_1[Y_i])\right\|_2,
        \left\|A_2[Y_i] -  
            f_{IPI}(A_2[Y_i])\right\|_2
    )
    \label{eqn:symm-loss-g}
\end{equation}

In addition to image loss, we also added an L1-regularization term
for image shift to avoid unrealistic large shift predictions
and keep the reconstructed 3D object near the origin of the coordinate system.
\begin{equation}
\small
    \mathcal{L}_{IPI} := \mathcal{L}_{\text{symm}}^{G} + 
    \frac{1}{2B}\sum_{i=1}^{B}\left\|t_i^{pred}\right\|_1
    \label{eqn:ipi-loss}
\end{equation}
\vspace{-10pt}
\paragraph{\IPI{} warm-up labeling} \label{sec: warm-up}
High-frequency features are difficult to learn in at an
early training stage, especially for projection images with
low SNR. We found that the training results can be improved
by using the low-pass filtered input images as training labels instead of using  original image labels,
as a training warm-up. 
The training image label ${\tilde{Y_i}}$ is defined below.
The details of $f_{\text{filter-k}}$ can be found in \SI{} \ref{Encoder and decoder}.
$f_{\text{filter-1}}$ is the first Gaussian filter with the 
lowest Gaussian convolution kernel variance,
while $f_{\text{filter-k}}$ is the k-th Gaussian filter with 
higher convolution kernel variance.
$N_{\text{warm-up}}^{IPI}$ is the iteration threshold
for switching to the original input images.
In practice, 3, 4 or 5 is selected as $k$ value.
\vspace{-5pt}
\begin{equation}
\small
    \tilde{Y_i} = \gamma (Y_i)
    +(1-\gamma)f_{\text{filter-k}}(Y_i)
    \label{eqn:IPI-warmup-gt-image}
\end{equation}
\vspace{-5pt}
\begin{equation}
\small
    \gamma = 
    \begin{cases} 
        0, & \text{if iteration}\, < N_{\text{warm-up}}^{IPI} \\ 
        1, & \text{if iteration}\, >= N_{\text{warm-up}}^{IPI} \\
    \end{cases}
    \label{eqn:IPI-warmup-schedule}
\end{equation}

\vspace{-8pt}
\subsection{\PIP{} (Pose-Image-Pose) task}
\vspace{-5pt}
The \PIP{} task is similar to the \IPI{} task but with 
\encoder{} and \decoder{} placed in reverse order. Besides, 
an additive Gaussian noise 
$\epsilon \sim \mathcal{N}( \mu, \sigma^2 I)$ is added to the 
output image from \decoder{} to create more realistic
inputs for the \encoder{}. The $\mu$ and $\sigma$ of the Gaussian noise
is sampled based on the background of the input images or on a given SNR.
Compared to the \IPI{} task, the inputs of \PIP{}
have changed from images to synthesized projection parameters 
$(R_i, t_i)$ drawn from certain distributions.
The loss function is defined as the MSE loss of 
rotation matrices $\{R_i\}_{i\in {\mathcal{B}}}$ and translation vectors 
$\{t_i\}_{i \in \mathcal{B}}$ for each batch $\mathcal{B}$ of size $B$ as shown below,
where 
$\left\| \cdot \right\|_F$ is the Frobenius matrix norm.
\begin{equation}
\small
    PIP: (R_i^{syn}, t_i^{syn}) \mapsto (R_i^{pred}, t_i^{pred})
    \label{eqn:pip-mapping}
\end{equation}
\vspace{-5pt}
\begin{equation}
\small
    f_{PIP}(R_i^{syn}, t_i^{syn}) := E_{IP} (D_{PI}(R_i^{syn}, t_i^{syn}) + \epsilon)
    \label{eqn:pip-ef}
\end{equation}
\vspace{-5pt}
\begin{equation}
\small
    \mathcal{L}_{PIP} = \frac{1}{B} \sum_{i=1}^{B}
        ( 
            \frac{1}{9}\left\|R_i^{syn} - R_i^{pred}\right\|_2 +
            \frac{1}{2}\left\|t_i^{syn} - t_i^{pred}\right\|_1
        )
\end{equation}
With only \IPI{}, the pose estimation of \encoder{} is not distributed in the whole pose space,
and some part of the pose space is never reached.
By applying \PIP{}, in which the input pose is randomly sampled in the whole pose space,
\encoder{} is forced to match the image-pose pairs over all possible pose spaces.
Therefore, \PIP{} gives \encoder{} a proper output distribution and lets the model measure infinity of pose and images instead of only images in datasets.
With a substantially correct volume representation of \decoder{},
\PIP{} even turns the unsupervised task into a supervised image-pose matching task.
When training the \PIP{} task, one has the choice of freezing 
the \decoder{} parameters and only training the \encoder{}, or allowing
both the \encoder{} and \decoder{} parameters to be updated.\looseness=-1

\vspace{-8pt}
\subsection{Training of the \IPI{} and \PIP{} tasks}
\vspace{-6pt}
The \IPI{} and \PIP{} tasks can be trained together or in succession.
When trained together, we designed the following training schedule with a warm-up
period of $N_{\text{warm-up}}^{\text{train}}$ iterations, 
which allows the \IPI{} task to learn an approximate description of the underlying
3D object before adding the \PIP{} task. 
Although we found the following simple schedule was sufficient in our benchmark tests,
other schedules of $\beta$ are also possible.
\vspace{0pt}
\begin{equation}
\small
    \mathcal{L}_{total} = \mathcal{L}_{IPI} + \beta \mathcal{L}_{PIP}
    \label{eqn:total-loss}
\end{equation}
\vspace{-10pt}
\begin{equation}
\small
        \begin{cases}
            \beta = 0, & \text{if iteration}\, < N_{\text{warm-up}}^{\text{train}} \\ 
            \beta >0, & \text{if iteration}\, >= N_{\text{warm-up}}^{\text{train}} \\
        \end{cases}
\end{equation}

\vspace{-10pt}
\section{Results}
\vspace{-6pt}

To evaluate the \textit{ab initio} 3D reconstruction quality of ACE-EM, 
we benchmarked this method against both the traditional cryo-EM 
3D reconstruction software cryoSPARC \citep{Punjani_2017_NatMeth} 
and recent methods,
including cryoPoseNet \citep{Nashed_2021_ICCV}, 
cryoDRGN2 \citep{Zhong_2021_ICCV}, 
and cryoAI \citep{Levy_2022_ECCV}, 
on both simulated and experimental cryo-EM datasets.

\vspace{-6pt}
\subsection{Dataset Preparation and Training Setup}
\label{Dataset Preparation and Training Setup}

\vspace{-6pt}
\paragraph{Datasets} Both simulated and experimental cryo-EM datasets are prepared 
following the setup of cryoAI \citep{Levy_2022_ECCV}.
For simulated datasets, 
two protein molecules,
the pre-catalytic spliceosome and 
the SARS-CoV-2 spike ectodomain structure,
are selected as the reconstructed targets.
The simulated dataset generation method and the detailed dataset setting can be seen in the \SI{} \ref{sec:Benchmark dataset}.
The shape of each projection image is 
${128\times128}$.
Each reconstructed volume is a voxel with 
${128\times128\times128}$ shape.
The experimental cryo-EM dataset
80S ribosome (EMPIAR-10028) consists of 
105,247 images with a shape of $360\times360$,
which are downsampled to $128\times128$ for training with \name{}.
Note that as discussed in the original publication \citep{Levy_2022_ECCV},
the cryoAI algorithm
could not handle 
$128\times128$-downsampled images due to 
convergence issues and an input image size of 
$256\times256$ was used during training.
The output shape of the reconstructed volume from cryoAI was set to
 $128\times 128\times128$ voxels,
the same as
\name{} and other methods.

\vspace{-10pt}
\paragraph{Accuracy assessment} The benchmark metrics follow cryoAI \citep{Levy_2022_ECCV}.
3D reconstruction quality was assessed by the Fourier Shell Correlation (FSC). 
It measures the correlation at every corresponding frequency or resolution(reciprocal of the frequency) between two volumes.
The FSC resolution at value ${k}$ is defined as
the specific resolution when the correlation is dropped to ${k}$.
  Rotation matrix error (Rot.) was calculated using 
  the mean/median square Frobenius norm relative to 
  ground-truth $R_i$. 
translation vector error (Trans.) was calculated by the mean square L2-norm of the predicted and ground-truth translation vector.

\vspace{-10pt}
\paragraph{Training setup} We ran our benchmark tests on a server 
with 8 Nvidia V100 GPUs and 84 CPU cores.
When training with the \FNet{} decoder, we used 
the batch size of 384 and the learning rate of 1e-4 
for both the encoder and decoder. \FNet{} was trained for 40,000 iterations.
When training with the \VGrid{} decoder, 
we used the batch-size of 1,024 and training iterations of 30,000. The learning rate 
for the encoder was 1e-3 and 1.0 for the decoder.
The duration of IPI warm-up labeling ${N_{\text{warm-up}}^{IPI}}$ usually set to 0 or 3000-8000 iterations.
The``WarmupMultiStepLR'' warm-up schedule and AdamW\citep{DBLP:journals/corr/abs-1711-05101} optimizer were used 
in both cases.
$N_{\text{warm-up}}^{\text{train}}$ is usually set to 0 or 2000-3000 iterations,
PIP parameter $\beta$ is usually set to 0.9 for 200 dB dataset and 0.05-0.09 for -10db dataset in both cases. 
The decoder is frozen during the \PIP{} process.

\vspace{-6pt}
\subsection{3D reconstruction on simulated datasets}

\vspace{-6pt}
\paragraph{Dataset} Two protein molecules are selected as the targets:
the pre-catalytic spliceosome (PDB ID: 5NRL) and 
the SARS-CoV-2 spike ectodomain structure (PDB ID: 6VYB). 
For comparing the performances of methods with different noise levels, 
we synthesized each dataset with Gaussian noises of different SNRs, 
200 dB, -10 dB or even -20 dB(Figure~\ref{fig:2dimgs}). 
With high SNR at 200 dB, images can be regarded as noise-free. 
-10 dB is the most common SNR in experimental projection image datasets, 
thus, the model performance of this condition is a focus. 
With -20 dB, the images become indistinguishable from human eyes 
and reconstruction becomes more challenging.
An additional dataset spliceosome at the SNR of -20 dB 
is applied to test the noise tolerance of our method.

\vspace{-10pt}
\paragraph{Baselines} We benchmarked against CryoPoseNet \citep{Nashed_2021_ICCV}, 
CryoSPARC \citep{Punjani_2017_NatMeth} (traditional method), cryoDRGN2 \citep{Zhong_2021_ICCV},
and cryoAI \citep{Levy_2022_ECCV} in terms of pose estimation
and 3D reconstruction accuracy.
For simulated datasets, FSC resolution at a threshold of 0.5 between the ground-truth and predicted volumes is the evaluation criterion.

\vspace{-10pt}
\paragraph{Results} The result of the simulated datasets can be seen in Table~\ref{table:benchmark}.
Results for CryoPoseNet were taken from previously reported data \citep{Levy_2022_ECCV},
cryoDRGN2's results were from \citep{Zhong_2021_ICCV}, while the cryoSPARC and cryoAI's results were obtained by re-running these methods.
We found that cryoAI failed frequently due to the spurious structural
symmetries even with symmetry loss applied, while \name{} rarely failed in reconstructions. 
The pose estimation accuracy and 3D reconstruction quality 
are comparable or better than these baseline methods using either \VGrid{} or \FNet{} decoder.
%\looseness=-1
In terms of reconstruction resolution, 
\name{} with \FNet{} outperformed all the baseline methods. 
It reached 2 pixels, 
the Nyquist resolution, 
using an FSC threshold of 0.5 in all datasets with SNR 200 dB and -10 dB.
Visualization of 3D reconstruction results can be found in Figure~\ref{fig:6vyb-200dB-volume} and Figure~\ref{fig:5nrl-10dB-volume}. Our reconstructed volumes have more details compared to the result of cryoAI. 
Based on the Nyquist–Shannon sampling theorem~\citep{Shannon1949}, 
the highest measurable resolution of the dispersed volume representation is 2 pixels. 
The resolution of \name{} with \FNet{} reached the theoretical upper limit,
which is not reachable by other methods.
To be specific, 
the FSC correlation coefficient at spatial resolution 2.00 pixels is still very high (0.86) for the spliceosome (200 dB) dataset,
well above the standard threshold (0.5).
We also benchmarked \name{} on an even more challenging dataset
(SNR -20 dB; without image shifts).
The FSC resolution of the output 3D object reached 2.1 pixels,
which is even higher than the results of -10 dB provided by other ML-based baselines.
Also, the performance of -20 dB SNR proves the potential of \name{} with \FNet{}
for working in really low SNR situations.

When using \VGrid{} as the decoder, 
\name{} reached a similar or better FSC resolution 
among other baselines.
In detail, on the similar SNR -10dB with the experimental environment,
it provided higher resolutions on the spliceosome dataset compared to other baselines (except \name{} with \FNet{}).
With \VGrid{}, CryoPoseNet failed in the spike datasets,
indicating that our method recovered \VGrid{}'s  competitiveness for cryo-EM 3D reconstructions.

For rotation errors and translation errors, our methods get comparable or better results with other baselines.
\name{} generated fairly accurate pose estimations on noise-free datasets as most baselines.
Moreover, \name{} performed better on noisy datasets compared to other ML-based methods, especially in terms of the mean rotation error metric.

\vspace{-6pt}
\subsection{3D reconstruction on experimental datasets}
% \vspace{-8pt}

\paragraph{Dataset} 80S EMPIAR-10028 is an experimental cryo-EM reconstruction dataset used for comparing baselines. The size is downsampled to 128 for cryoDRGN2 and our methods, and it is downsampled to 256 for cryoSPARC and cryoAI due to the failure of reconstruction with cryoAI~\citep{Levy_2022_ECCV} in size 128.

\vspace{-10pt}
\paragraph{Baselines} For experimental datasets, 
evaluation criterion is half-map FSC~\citep{ROSENTHAL2015135} resolution at 0.143, 
which splits the dataset evenly into two half datasets, 
then trains the two datasets separately to get two volumes for comparison.

\begin{table*}[t]
\vspace{-10pt}
  \caption{Accuracy of pose estimation and 3D 
  reconstruction using per-image pose search
  (cryoSPARC and cryoDRGN2) and amortized inference 
  (cryoPoseNet, cryoAI, and \name{}). 
  Resolution(Res.; unit: pixels), rotation matrix error (Rot.) and 
  translation vector error(Trans.) are listed.
  \name{} with \VGrid{} is shown as ``ours (V)''
  and \name{} with \FNet{} decoder as ``ours (F)''.
  The best results are in bold and the second-best results are underlined.
  }
  \label{table:benchmark}
  \vskip 0.15in
  \begin{center}
    \begin{small}
      \begin{sc}
  \addtolength{\tabcolsep}{-2.5pt}
  \begin{tabular}{ll|l|l|lllll}
    \toprule
            &&  \multicolumn{1}{c|}{traditional} & \multicolumn{5}{c}{ML-based}\\
    \cmidrule{3-8}
    Dataset  &&  cryoSPARC & cryoDRGN2 & cryoPoseNet & cryoAI & ours (V) & ours (F)\\
    \cmidrule{3-8}
     &&  \multicolumn{2}{c|}{per-image pose search} & \multicolumn{4}{c}{amortized inference}\\

    \midrule
    spliceosome & Res.  & \underline{2.06} & -&  2.78 & 2.16 & 2.23 & \textbf{2.00}\\
     $\infty$/200dB & Mean Rot. & 0.01 & - & - & \underline{0.0007} & 0.004 & \textbf{0.0004} \\
     & Med Rot.  & \textbf{0.0003} & - & 0.004 & \underline{0.0005} &  0.0007 & \textbf{0.0003} \\
     & Trans. &  \underline{1.2} & - & - & \textbf{0.9} & 10.3 & 1.7 \\
    \midrule
    spliceosome  & Res. &  \underline{2.06} & - & 3.15 & 2.65 & \textbf{2.00} & \textbf{2.00}\\
     -10dB & Mean Rot. & 0.01 & - & - & 0.03 & \underline{0.005}  & \textbf{0.003} \\
     & Med Rot. & \textbf{0.00006} &  - & 0.01 & 0.003 & 0.002  & \underline{0.001} \\
     & Trans.  & \textbf{1.2} & - & - & 3.2 & \underline{1.7} &2.1 \\
    \midrule
    spliceosome  & Res. & 2.24 & - & - & - & - & \textbf{2.10} \\
     -20dB & Mean Rot. & \textbf{0.6} & - & - & - & - & 1.3 \\
     & Med Rot. & \textbf{0.0007} & - & - & - & - & 0.02 \\
     & Trans. & 11.8 & - & - & - & - & -\\
    \midrule
    spike &  Res.  & \underline{2.06} &  - & 16.0 & 2.12 & 2.85 & \textbf{2.00}\\
      $\infty$/200dB & Mean Rot. & 0.007   & \underline{0.0004} & - & 0.0006 & 0.05 & \textbf{0.0003} \\
      & Med Rot.  & 0.0007  & \textbf{0.0001} & 5 & 0.0004 & 0.002 & \underline{0.0002} \\
     & Trans.  & \textbf{0.03} & - & - & \underline{0.7} & \underline{0.7} &1.2\\
    \midrule
    spike  & Res.  & 2.06 &  \underline{2.03} & 16.0 & 2.28 &2.91 &\textbf{2.00}\\
      -10dB & Mean Rot. & \textbf{0.0005} &  0.06 & - & 0.06 & 0.006 & \underline{0.003}\\
     & Med Rot.  & \textbf{0.0002} &  0.01 & 6 & 0.001 & 0.002 & \underline{0.0006} \\
     & Trans.  &  \textbf{0.03} & - & - & 0.8 & \underline{0.4} &1.3\\
    \bottomrule
  \end{tabular}
    \end{sc}
  \end{small}
\end{center}
\vskip 0in
\end{table*}

\vspace{-10pt}
\paragraph{Results} The result of the experimental datasets can be see in Table~\ref{table:benchmark_exp}.
Results for cryoSPARC, cryoDRGN2 and cryoAI were taken from
previously reported data \citep{Levy_2022_ECCV}, 
The input image size and output volume size setting 
for different methods can also be seen in Table~\ref{table:benchmark_exp}.
CryoAI claimed it is the first amortized inference method to demonstrate proper volume reconstruction on an
experimental dataset. 
However, CryoAI did not properly converge when
fed with input images of size 128. 
A compromise is reached
by training with input images with a 256-pixel side but producing a 128-pixel side volume in cryoAI.
CryoAI is a standard auto-encoder architecture with \FNet{} as its decoder. 
Therefore, using \name{} with \FNet{} is almost like adding the PIP task to the CryoAI structure.
The convergence problem is preliminarily resolved by applying our method.
After adding the PIP task, 
the model can adequately converge with input size ${128\times128}$.
Also, the resolution of \name{} with \FNet{} on a 128-pixel input size is better than CryoAI on a 256-pixel input size, 
which shows the robustness and effectiveness of our method.
Meanwhile, the result of \name{} with \VGrid{} is on the Nyquist resolution, representing its capacity for the experimental situation.
Visualization of 3D reconstruction result can be seen in Figure~\ref{fig:6vyb-200dB-volume}.

\begin{table*}[t]
  % \vspace{-10pt}
  \caption{Accuracy of 
  3D 
  objection reconstruction for experimental 80S ribosome dataset.
  Resolutions (Res.) are shown in angstroms 
  (Nyquist resolution: 7.54 \AA{}).
  \name{} with \VGrid{} and \FNet{} are labeled ``ours (V)''
  and ``ours (F)'', respectively.
  }
  \label{table:benchmark_exp}
  \vskip 0.15in
  \begin{center}
    \begin{small}
      \begin{sc}
  \addtolength{\tabcolsep}{-1pt}
  \begin{tabular}{l|llllllll}
    \toprule
     &   \multicolumn{1}{c|}{traditional} & \multicolumn{5}{c}{ML-based}\\
    \cmidrule{2-7}
    80S ribosome (exp.) & cryoSPARC & cryoDRGN2 & cryoAI & cryoAI & ours (V) & ours (F)\\
    \cmidrule{2-7}
    & \multicolumn{2}{c|}{per-image pose search} & \multicolumn{4}{c}{amortized inference} \\
    \midrule
    Input image side size & 256 & 128 & 128 & 256 & 128 & 128 \\
    Output volume side size & 256 & 128 & 128 & 128 & 128 & 128 \\
    Final volume side size & 128 & 128 & 128 & 128 & 128 & 128 \\
    \midrule

    Res. \AA($\downarrow$) & 7.54 & 7.54 & Fail & 7.91 & 7.54 & 7.66  \\
    \bottomrule
  \end{tabular}
    \end{sc}
  \end{small}
\end{center}
% \vskip -0.2in
\end{table*}

\vspace{-5pt}
\subsection{Ablation studies}
\label{sec:Ablation studies}
\vspace{-5pt}
Without \PIP{}, \name{} is degenerate into a traditional autoencoder structure. 
Comparisons are made between training with and without
\PIP{} tasks in two different decoders,
to prove that our work \name{} can boost the performance of an autoencoder by applying \PIP{} regardless of decoder types. \looseness=-1

\vspace{-6pt}
\subsubsection{\PIP{} ablation with \FNet{}}
\begin{figure*}[t] 
  \centering
\begin{minipage}[t]{0.329\linewidth}
  \centering
  \includegraphics[width=1\linewidth]{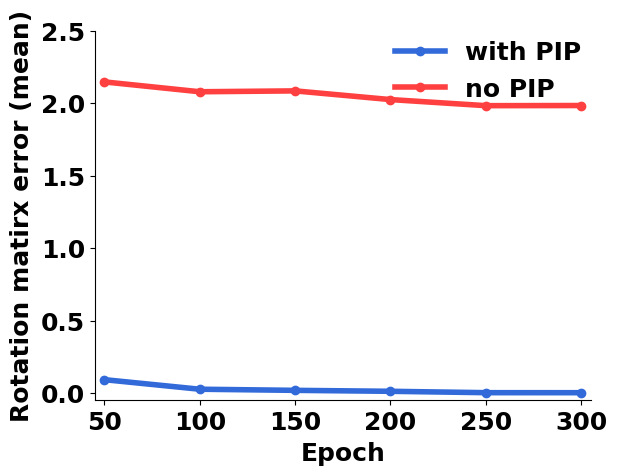}
\end{minipage}
\begin{minipage}[t]{0.329\linewidth}
  \centering
  \includegraphics[width=1\linewidth]{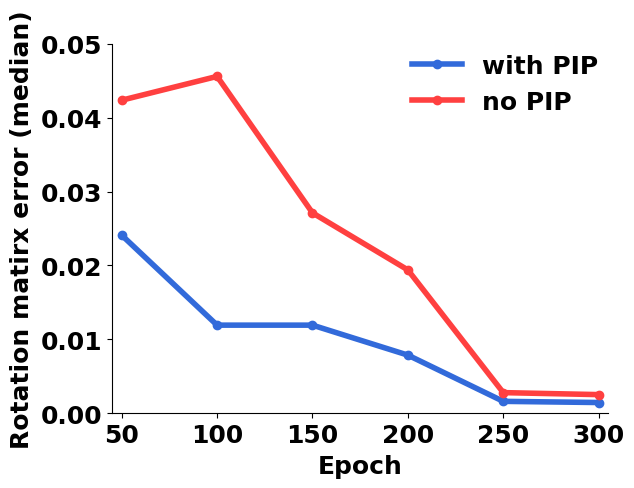}
  \
\end{minipage}
\begin{minipage}[t]{0.329\linewidth}
  \centering
  \includegraphics[width=1\linewidth]
      {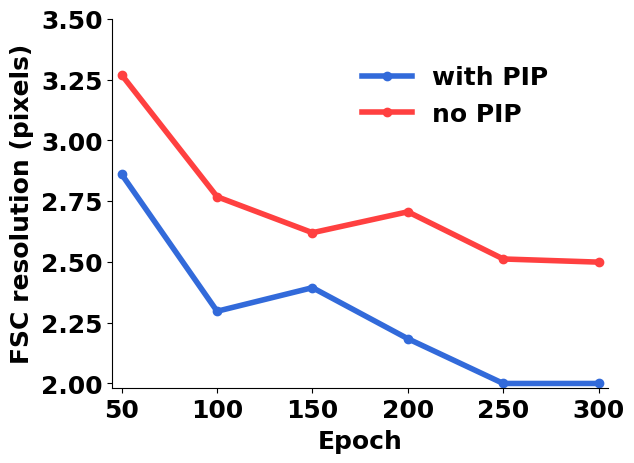}
  \
\end{minipage}
\vspace{-20pt}
  \caption{Ablation study with (blue) or without (red) the \PIP{} task 
    using the spliceosome dataset (-10 dB)
    with \FNet{} as the decoder.
    \textbf{Left:} mean rotation matrix prediction error.
    \textbf{Middle:} median rotation matrix prediction error.
              \textbf{Right:} convergence of the FSC-resolution.
  }
\label{fig:ablation-pip-rotmat-error}
\end{figure*}

\vspace{-8pt}
To test the effect of \PIP{} with \FNet{},
we ran an ablation study on the spliceosome dataset (-10 dB) 
using a batch size of 384 for 300 epochs.
As shown in Figure~\ref{fig:ablation-pip-rotmat-error},
all the performances, including FSC-resolution, mean and median rotation matrix prediction errors, are much worse without the \PIP{} task.
Though the median rotation error is similar at the end, 
the gap in the mean rotation error is noticeable. 
Without \PIP{} task, the mean rotation error is 1.98 at the 300th epoch, 
which is three orders of magnitude larger than 0.003, the mean error of \name{}.
\PIP{} reduces the existence of exaggerated rotation errors(see in Figure \ref{fig:rotmat-error-hist}), 
then gets a better mean rotation error and further improves the resolution.

We also compare the coverage of the projection orientation space
during training in terms of the distribution of the
oriented z-axis 
(projection direction), which is equivalent to 
the last column of the predicted rotation matrix.
When creating the simulated datasets, 
the projection orientations were sampled from 
a uniform distribution.
Therefore the directions of the oriented z-axis should also be uniformly distributed.
As shown in Figure~\ref{fig:ablation-z-ball},
the predicted projection direction vectors covered 
the entire orientation space with the \PIP{} task 
starting from epoch 50. 
In contrast, without the \PIP{} task, 
the coverage of the orientation space is still incomplete 
at around 20\% in 300 epochs.
And the insufficient coverage performance reflected in the large gap between the mean and median rotation errors.
After adding \PIP{} task, the training time is increased to 1.25-1.50 times of the original duration for one epoch.
But the time consumption is worthy of faster convergence and better performances.

\vspace{-8pt}
\subsubsection{\PIP{} ablation with \VGrid{}}
\begin{figure}[htbp] 
	\begin{minipage}[t]{0.49 \textwidth}
		\raggedright
		\includegraphics[width=1\linewidth]{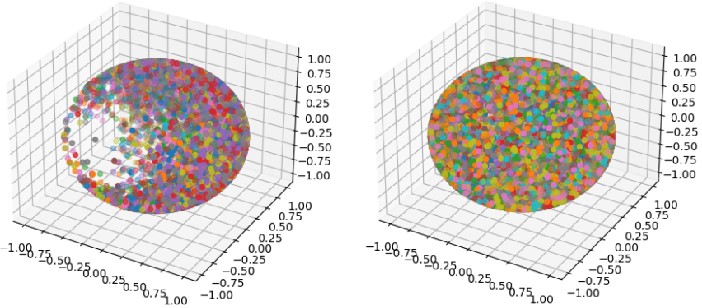}
    	\caption{
    		Visualization of the predicted projection directions for the spliceosome dataset (-10 dB).
    		\textbf{Left}:  without \PIP{} (epoch 300);
    		\textbf{Right}: with \PIP{} (epoch 50).
    	}
  		\label{fig:ablation-z-ball}
	\end{minipage}
	\hfill
	\begin{minipage}[t]{0.49 \textwidth}
		\raggedleft
		\includegraphics[width=1\linewidth]{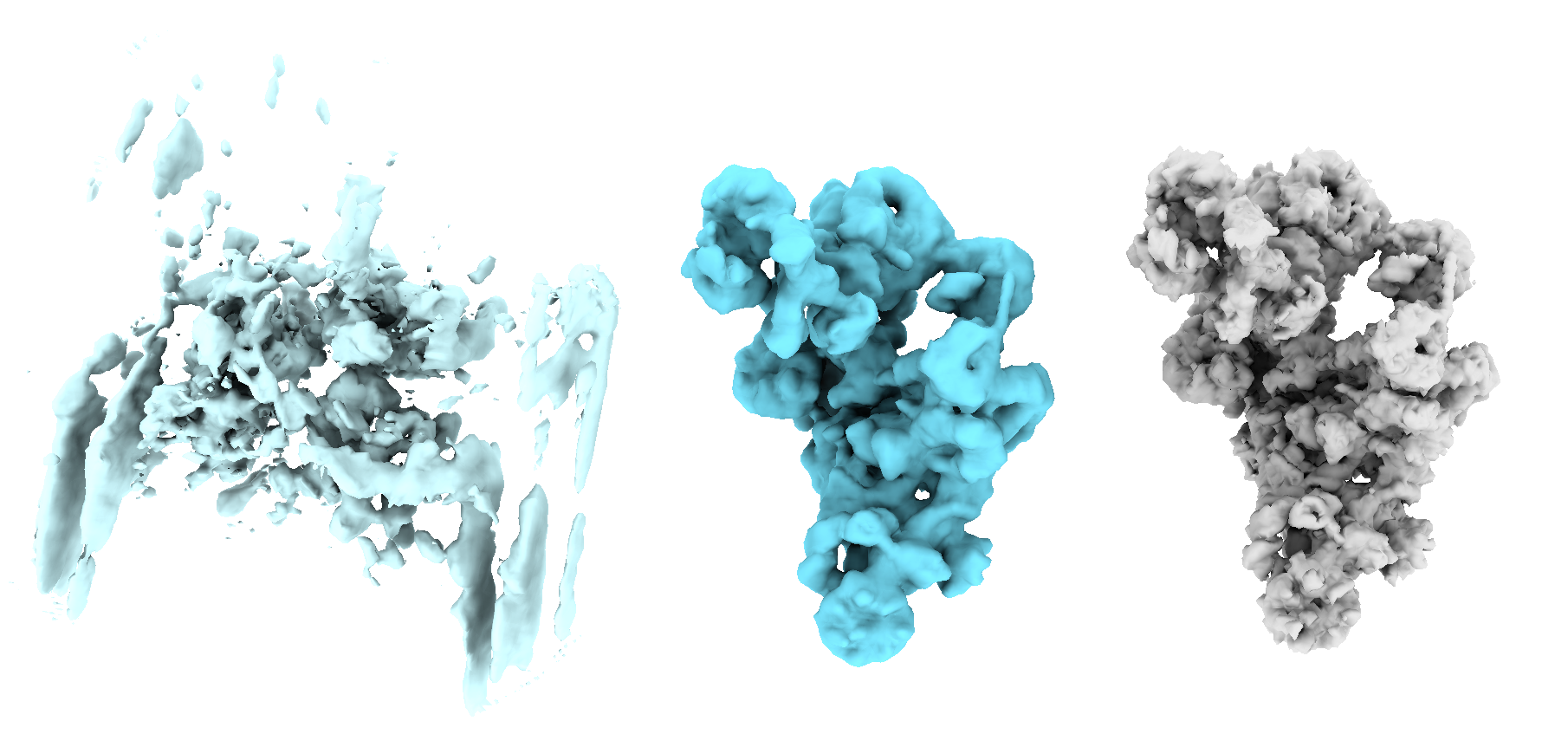}
	
    	\caption{
    			3D reconstruction of the spliceosome dataset (-10 dB) 
    			with \VGrid{} without 
    			(\textbf{left}) or with (\textbf{middle}) 
    			the \PIP{} task. The ground-truth is shown on the \textbf{right}.
    	}
            \label{fig:ablation-pip-voxel}
	\end{minipage}
\end{figure}

\vspace{-8pt}
We also perform an ablation study for the \PIP{} task using 
\VGrid{} as the decoder on the same spliceosome dataset (-10 dB) using a batch size of 1,024 for 300 epochs.
As shown in Table~\ref{table:ablation-pip-voxel}
and Figure~\ref{fig:ablation-pip-voxel},
3D reconstruction failed without either the \PIP{} task or warm-up labeling (Section \ref{sec: warm-up}).
The reconstruction can be completed when using 
warm-up labeling for loss calculations
at the beginning of training.
When \PIP{} task was included, the mean rotation matrix prediction error was reduced significantly, and the FSC resolution was also improved. 

\vspace{40pt}

\vspace{-5pt}
\section{Conclusion}
\vspace{-6pt}
In this work, we have developed a new unsupervised 
learning framework ACE and applied it to the 
\textit{ab initio} cryo-EM 3D reconstruction problem.
Compared to existing methods, the most significant 
advantage of \name{} is its ability to learn 
the pose space much more effectively and accurately. 
The \name{} method can work with different types of decoders,
and boost the FSC-resolution even to Nyquist resolution
in some simulated and experimental datasets.
In the future work on \name{}, 
we plan to extend the ACE framework to work with 
heterogeneous cryo-EM 3D reconstruction and 
novel view synthesis for natural images.

% \newpage
\section{Reproducibility Statement}
The model structure can be seen in Section \ref{Approach} and \SI{} \ref{Encoder and decoder}.
The datasets and training setup can be seen in Section \ref{Dataset Preparation and Training Setup}.
The simulated dataset generation method and the detailed dataset setting can be seen in the \SI{} \ref{sec:Benchmark dataset}.
The code to reproduce our experiments will be open-sourced upon publication.

%%%%%%%%% REFERENCES
{\small
\bibliographystyle{unsrt}
\bibliography{ace_refs}
}

\newpage
\newpage
\appendix
\onecolumn

\section{Encoder and decoder}
\label{Encoder and decoder}
\subsection{Encoder \encoder{}}
The \encoder{} represents a function that maps an input image $Y_i$
to its corresponding projection pose parameters $(R_i, t_i)$.
$R_i \in SO(3) \subset \mathbb{R}^{3\times3}$ is a rotation matrix
for mapping a reference orientation to the projection orientation.
$t_i \in \mathbb{R}^2$ is the 2D translation vector to
account for the 2D image shift in the input image $Y_i$.

\begin{equation}
E_{IP}: Y_i \mapsto (R_i, t_i)
\label{eqn:encoder-def1}
\end{equation}

Cryo-EM projection images are extremely noisy. 
To improve the \encoder{} performance, 
we adopted the same image preprocessing strategy as 
in cryoAI \citep{Levy_2022_ECCV}.
Each image $Y_i$ is passed on to a set of five
Gaussian filters,
which are implemented as 2D convolutions
with a kernel size of 11 pixels and output a 5-channel image $Y_i^{5C}$.
The Gaussian kernel variances follow a geometric series,
i.e., $10^{-2}$, $10^{-1}$, $1$, $10$, $10^2$ in $\text{pixel}^2$
for output image channels 1 to 5. The filtered images 
are given to a ResNet-18 \citep{He_2016_CVPR} network with two 
MLP networks for the inference of $R_i$ and $t_i$ respectively.
The raw prediction of $t_i$ is passed through a Sigmoid function
to restrict the range of values to $(0, 1)$, because $t_i$ 
is expressed in unit of fraction of the image edge length.
Other feature extractors (FE) can also be used in place of ResNet-18.
\begin{equation}
    f_{filter}: Y_i \mapsto Y_i^{5C}
    \label{eqn:filter-def}
\end{equation}

The complete \encoder{} can be defined as follows,
where ``$\circ$" means function composition. the trainable parameters
are the network work parameters in $f_{FE}$ and $f_{MLP}$.
\begin{equation}
    E_{IP}(Y_i) := (f_{MLP} \circ f_{FE} \circ f_{filter})(Y_i)
    \label{eqn:encoder-def2}
\end{equation}

There are many representations of $R_i$ such as Euler angles
and quaternions. It has been shown that all 3D rotation
representations in the real Euclidean spaces with no more than
four dimensions are discontinuous and 
difficult to learn using neural networks \citep{Zhou_2019_CVPR}.
We chose the 6-dimensional vector representation 
in $\mathbb{R}^3$ \citep{Zhou_2019_CVPR}, which can be
converted into a rotation matrix in $\mathbb{R}^{3\times3}$
using PyTorch3D \citep{Johnson_2020_SIGGRAPH}.

\subsection{Decoder \decoder{}}
The \decoder{} takes pose $(R_i, t_i)$ and 
 predicts the projection image $Y_i^{pred}$.

\begin{equation}
    D_{PI}: (R_i, t_i) \mapsto Y_i^{pred}
    \label{eqn:decoder-def1}
\end{equation}

There are many possible choices for \decoder{}. 
we tested a real-space voxel grid
decoder \VGrid{} which was 
used in cryoGAN~\citep{Gupta_2021_IEEE} and partially used in CryoPoseNet~\citep{Nashed_2021_ICCV},
and a frequency-space neural network decoder 
\FNet{} 
from cryoAI~\citep{Levy_2022_ECCV}.
The trainable parameters of \decoder{} 
depend on the implementation.
In \VGrid{}, the trainable parameters
are stored in a 3D tensor $V \in \mathbb{R}^{L \times L\times L}$,
which is a discrete representation of 3D reconstruction target in real space.
In \FNet{}, the trainable parameters are the network parameters for 
the two Sinusoidal Representation Networks (SIRENs)~\citep{SIREN} as detailed in cryoAI~\citep{Levy_2022_ECCV}.

\section{Benchmark dataset}
\label{sec:Benchmark dataset}
To create the ground-truth volumes,
we first generate simulated cryo-EM density maps based on
published atomic structure coordinate (pdb) files using the 
molmap command from ChimeraX \citep{Petersen_2021_ProtSci}. 
In each density map, each atom is described by a 3D Gaussian 
distribution with a width proportional to a chosen 
``resolution" parameter
(6 \AA\, for both proteins).
Then, the ground-truth volumes were generated by 
re-sampling the simulated density maps onto cubic voxel grids
with 128 pixels on each side. 
A set of projection images were generated using cryoSPARC 
\citep{Punjani_2017_NatMeth} with
uniformly sampled projection orientations over the ground-truth volume.
To reduce the difference between the simulated projection images and the experimental ones, 
random translations and CTF were applied to each projection image.
Random translations were sampled
from a Gaussian distribution ($\mu=0$, $\sigma = 20$ \AA).
\begin{table*}[htbp]
    \caption{}
    \label{table:dataset}
    \centering
    \begin{tabular}{lllllll}
      \toprule
      & Dataset & L & N & \AA /pix. & Shift? & SNR(dB)\\
      \midrule
      Simulated & Spike & 128 & 50,000 & 3.00 & N & $\infty$\\
      & Spike & 128 & 50,000 & 3.00 & N & -10\\
      & Spliceosome & 128 & 50,000 & 4.25 & Y & $\infty$\\
      & Spliceosome & 128 & 50,000 & 4.25 & Y & -10\\
      & Spliceosome & 128 & 50,000 & 4.25 & N & -20\\
      \midrule
      Experimental & 80S EMPIAR-10028 (downsampled) & 128 & 105,247 & 3.77 & Y & NA\\
  
      \bottomrule
    \end{tabular}
  \end{table*}

  \section{Background}
  Cryo-EM is concerned with the inverse problem of inferring 
  the 3D object associated with a set of projection images 
  $\{Y_i\}_{i \in \mathcal{D}}$ where $\mathcal{D}$ is index 
  set for the image dataset. Here we briefly introduce the 
  image formation model in cryo-EM and relevant terminologies. 
  
  \subsection{Cryo-EM image formation model}
  Cryo-EM projection images are formed by collecting
  the electrons scattered by the atoms from 
  macromolecules (e.g., protein molecules) in the sample embedded in thin ice. 
  The raw cryo-EM data are usually referred to as micrographs, which consist of
  the 2D projections of hundreds of 3D objects.
  In a so-called ``particle picking'' step,
  the projection images of individual 3D objects 
  are cropped out of each micrograph and collected
  into a stack of projection images.
  
  The 3D reconstruction target object can be 
  defined as a function $V$ for mapping 
  a 3D coordinate to a real value.
  \begin{equation}
      V: \mathbb{R}^3 \mapsto \mathbb{R}
      \label{eqn:volume-fn-mapping}
  \end{equation}
  Each projection image is associated with a 
  3D object oriented in a certain direction.
  Here we focus on the homogenous 3D reconstruction problem 
  where the 3D objects corresponding to all
  the projection images are identical except for their
  orientations which can be modeled as a rotation matrix
  $R_i\in SO(3) \subset \mathbb{R}^{3\times 3}$.
  Each 3D objection can be defined as follows, where
  the 3D coordinate $\vec{r}=[x, y, z]^T$.
  \begin{equation}
      V_i := V(R_i \vec{r})
      \label{eqn:volume-fn-def}
  \end{equation}
  The projection image $X_i$ of object $V_i$ is 
  described as integration along the projection direction, usually along the z-axis.
  \begin{equation}
      X_i(x, y) := \int_{\infty}^{\infty} V(R_i \vec{r}) dz
      \label{eqn:clean-projection-def}
  \end{equation}
  
  Due to the wavelike nature of electrons and the magnetic
  lens systems in the electron microscopes,
  the projection image signals $X_i$ are corrupted by 
  optical interference effects which are modeled by the
  Point Spread Function (PSF) $f_{PSF}$ in real space or
  the Contrast Transfer Function (CTF) in Fourier space
  (see the next section for details).
  Furthermore, the cropping process 
  of projection images for individual 3D objects
  from large micrographs is imperfect which
  can result in
  small translational image shifts.
  Lastly, cryo-EM images are affected by noise from
  various sources \citep{Baxter_2009_JSB} which is 
  assumed to be zero mean, uncorrelated, independently
  distributed \citep{Penczek_2010_MethEnzymol}.
  A common practice is to model it as a Gaussian noise 
  $\epsilon_i \sim \mathcal{N}(0, \sigma^2 I)$.
  Taken together,  the entire cryo-EM image formation model 
  can be described by the following equation 
  \citep{Levy_2022_ECCV,Bendory_2020_IEEE}
  where $Y_i$ is the final projection image considering
  all factors discussed above, 
  2D coordinate $\vec{p} = [x, y]^T$, and 
  2D translation $\vec{\tau} = [\Delta x, \Delta y]^T$.
  \begin{equation}
      Y_i(\vec{p}) = (f_{PSF} \ast X_i)(\vec{p} + \vec{\tau}) + \epsilon_i(\vec{p})
      \label{eqn:cryo-em-image-formation1}
  \end{equation}
  Since translation of a function by $\tau$ is equivalent to convolution with a shifted delta
  function: 
  \begin{equation*}
      \begin{aligned}
      f(x + \tau) = (\delta_\tau \ast f)(x)
      \text{, where } \delta_\tau := \delta(x + \tau)
      \end{aligned}
  \end{equation*}
  We can rewrite the above equation as below.
  \begin{equation}
      Y_i(\vec{p}) = (\delta_{\vec{\tau}} \ast f_{PSF} \ast X_i)(\vec{p}) + \epsilon_i(\vec{p})
      \label{eqn:cryo-em-image-formation2}
  \end{equation}
  The real-space VoxelGrid decoder used in this work was designed based on 
  this image formation model.
  
  \subsection{Fourier slice theorem}
  An alternative and more computationally efficient method of calculating image projection 
  is based on the Fourier slice theorem.
  It states that the Fourier transformation of 
  a projection image is the same as a slice of the Fourier transform
  of the corresponding 3D object. The orientation of this slice is the same
  as the projection plane.
  \begin{equation}
      \hat{Y}_i= \mathcal{S}_i[\hat{V}_i]
      \label{eqn:fourier-slice-theorem}
  \end{equation}
  Here $\hat{Y}_i$ and $\hat{V}_i$ are the Fourier transform of 
  the projection image $Y_i$ and $\hat{V}_i$, respectively.
  $S_i$ is the volume slicing operator which is defined as a mapping
  from frequency coordinates $\vec{k} = [k_x, k_y, k_z]^T$ on
  a rotated 2D Fourier plane (passing through the origin) 
  to the value of $\hat{V}_i$ at that point in Fourier space.
  \begin{equation}
      \hat{S}_i[\hat{V}_i](k_x, k_y) := \hat{V}(R_i \vec{k})\vert_{k_z=0}
      \label{eqn:slice-operator-mapping}
  \end{equation}
  Using the Fourier slice theorem, the cryo-EM image formation model
  in Fourier space can be simplified as follows,
  where 2D frequency coordinate $\vec{q} = [k_x, k_y]^T$.
  \begin{equation}
      \hat{Y}_i(\vec{q}) := 
          F_{\vec{\tau}}(\vec{q}) \cdot 
          F_{\text{CTF}}(\vec{q}) \cdot 
          S_i[\hat{V}] (\vec{q})
          +  \hat{\epsilon}(\vec{q})
      \label{eqn:em-image-formation-fourier}
  \end{equation}
  $F_{\text{CTF}}$ is the contrast transfer function which is the Fourier transform
  of the Point Spread Function \citep{Penczek_2010_MethEnzymol}. 
  $\hat{\epsilon}$ is the 
  Gaussian white noise in Fourier space. $F_{\vec{\tau}}$ is the phase-shift function defined below, which 
  is equivalent to translation by $\vec{\tau}$ in real space:
  $F_{\vec{\tau}}(\vec{q}) := e^{-2\pi i \vec{\tau}\cdot \vec{q}} $. 
  A majority of the cryo-EM 3D reconstruction algorithms are based on 
  this theorem, such as RELION \citep{Scheres_2012_JSB}, 
  cryoSPARC \citep{Punjani_2017_NatMeth}, 
  cryoDRGN \citep{Zhong_2019_ICLR,Zhong_2021_NatMeth,Zhong_2021_ICCV}, 
  and cryoAI \citep{Levy_2022_ECCV}.

\section{Visualization of 2D images and 3D reconstruction}
\begin{figure*}[htbp]
  \centering
      \includegraphics[width=1.\textwidth]{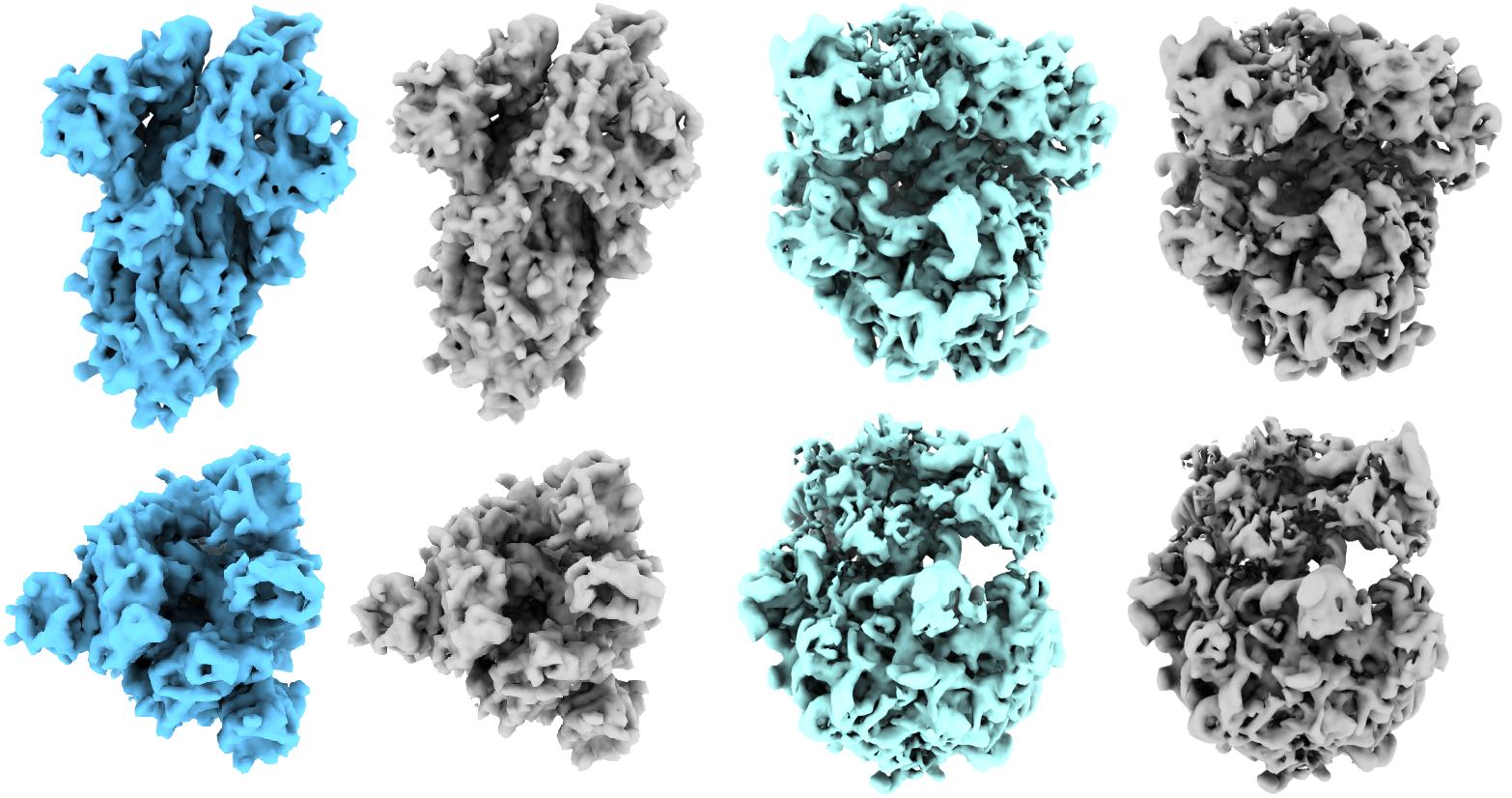}
      \caption{
        Visualization of the predicted 3D objects (in blue color) and the ground-truth (in gray color) 
        at two different viewing angles.
        \textbf{Left:} reconstruction of the spike (200 dB) dataset  using \FNet{}.
        \textbf{Right:} reconstruction of 80S ribosome using \VGrid{} with half of the experimental dataset. 
        The approximate ground-truth volume for 80S ribosome was constructed by 
        cryoSPARC using the entire set of experimental projection images (L = 360 pixels) and then
        downsampled to a smaller size (L = 128 pixels).
        }
  \label{fig:6vyb-200dB-volume}
\end{figure*}

\begin{figure*}[htbp]
  \centering
      \includegraphics[width=1.\textwidth]{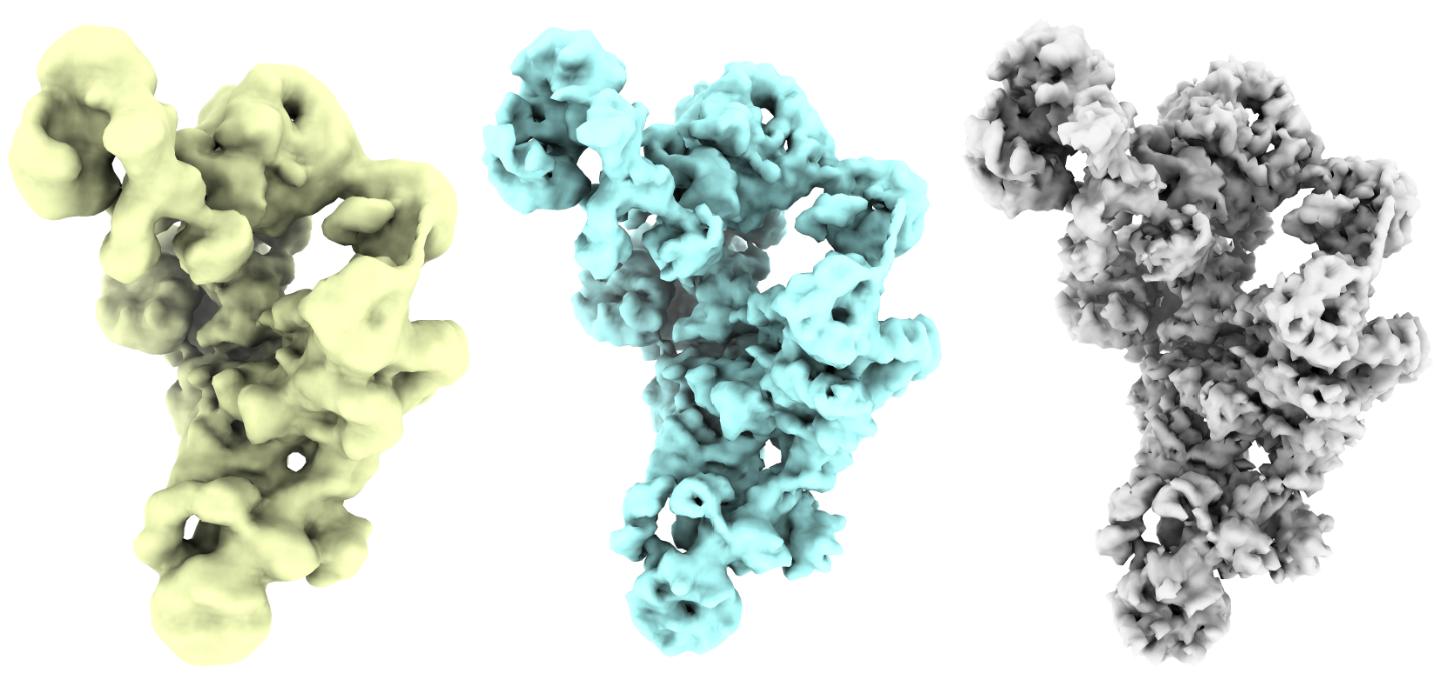}
      \caption{
        Visualization of the 3D objects of the spliceosome (-10dB).
        From left to right: reconstructions from cryoAI,  \name{} with \FNet{} and the ground-truth.
        }
  \label{fig:5nrl-10dB-volume}
\end{figure*}

Visulization of the 2D projection images
involved in the \name{} method is shown in \cref{fig:6vyb-200dB-volume}. 
An example of the 3D reconstruction results is shown 
in \cref{fig:5nrl-10dB-volume}.

\begin{figure*}[htbp]
  \centering
      \includegraphics[width=1.\textwidth]{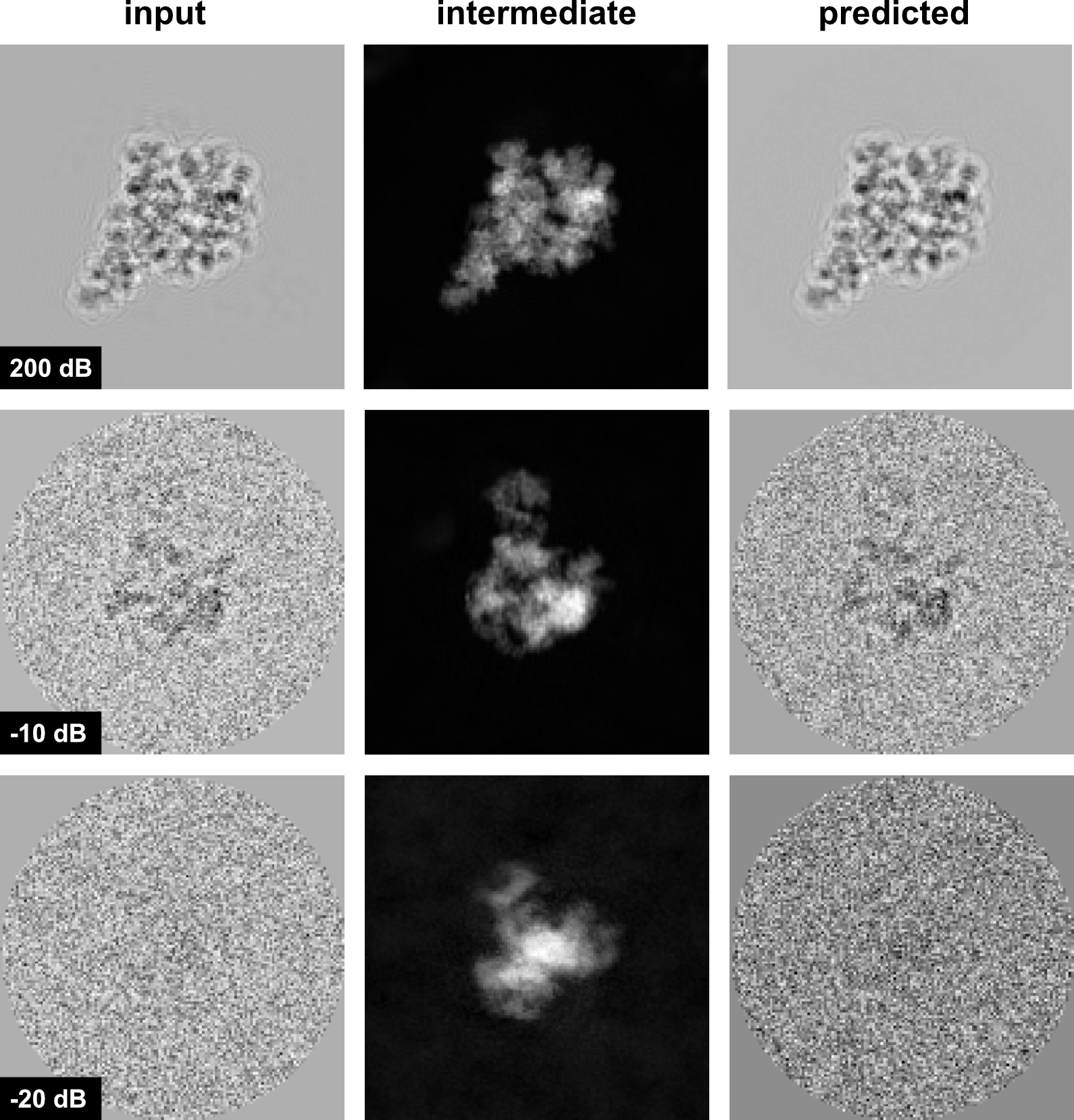}
      \caption{
        Visualization of the input and predicted 
        2D projection images by
        \name{} on the spliceosome datasets at different noise levels. 
        \textbf{From left to right:} The original images from the input datasets,
        the intermediate projection images generated by the decoder (\FNet{}),
        the final predicted images with CTF and noise applied.
        \textbf{From top to bottom:} datasets with SNR at 
        200 dB, -10 db, and -20 dB, respectively.
        }
  \label{fig:2dimgs}
\end{figure*}

\section{Ablation study of the PIP task}
\begin{figure}[htbp]
  \centering
      \includegraphics[width=0.75\textwidth]{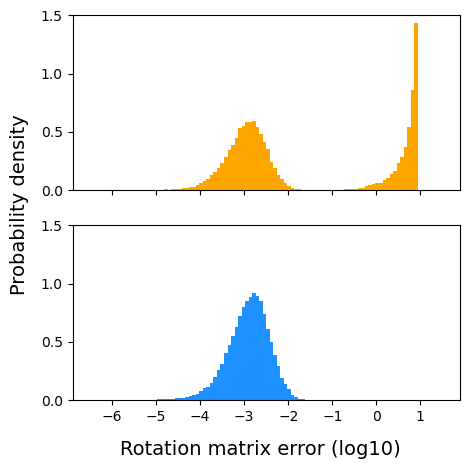}
      \caption{
        Distribution of rotation matrix prediction errors 
        for the spliceosome dataset (-10 dB).
        \textbf{Top:} without PIP;
        \textbf{Bottom:} with PIP.
        }
  \label{fig:rotmat-error-hist}
\end{figure}

Ablation study results for the PIP task are shown 
in \cref{table:ablation-pip-voxel},
\cref{fig:ablation-pip-rotmat-error}, 
and \cref{fig:rotmat-error-hist}.

\begin{table*}[htbp]
\vspace{-5pt}
  \caption{Ablation study for the \PIP{} task with \VGrid{}. ``Warm-up labeling'' refers to 
	using Gaussian-filtered input images as labels for loss calculation
	. FSC resolution (Res.) was calculated at 
	a threshold of 0.5
	against ground-truth volume (unit: pixels).
	Rotation errors (Rot.) are measured by mean and median values.
	}
  \label{table:ablation-pip-voxel}
  \centering
  \begin{tabular}{lllllllll}
    \toprule
    PIP task &  Warm-up labeling & Res. & Rot. (mean) & Rot. (median)\\
    \midrule
    No & No  & N/A & N/A & N/A \\
    No & Yes  & 2.24 & 1.124 & 0.005\\
    Yes & Yes  & 2.20 & 0.008 & 0.004\\
    \bottomrule
  \end{tabular}
\end{table*}

\end{document}